%% file: emnlp2024_camera_ready.tex
\pdfoutput=1

\documentclass[11pt]{article}

\usepackage[final]{acl}

\usepackage{times}
\usepackage{latexsym}

\usepackage[T1]{fontenc}

\usepackage[utf8]{inputenc}

\usepackage{microtype}

\usepackage{inconsolata}

\usepackage{graphicx}
\usepackage{enumitem}
\usepackage{amsmath}
\usepackage{amssymb}
\usepackage{mathtools}
\usepackage{amsthm}
\usepackage{graphicx}
\usepackage{xspace}
\usepackage{pifont}
\usepackage{array}
\usepackage{bm}
\usepackage{multirow}

\usepackage{rotating}
\usepackage{subcaption}
\usepackage{tablefootnote}
\usepackage{arydshln}
\usepackage{adjustbox}
\usepackage{wrapfig}
\usepackage{CJKutf8}
\usepackage{float}
\title{WavLLM: Towards Robust and Adaptive Speech Large Language Model}

\def\myand{\end{tabular}\hss\egroup \hfil\hfil\egroup
           \hbox to \linewidth\bgroup\large \hfil\hfil
             \hbox to 0pt\bgroup\hss \begin{tabular}[t]{c}\bf}
             
\author{
  Shujie Hu$^{1}$\thanks{Work was done during internship at Microsoft Research Asia.}~
  Long Zhou$^{2}$\thanks{Corresponding author.}~
  Shujie Liu$^{2}$~
  Sanyuan Chen$^{2}$~
  Lingwei Meng$^{1}$~
  Hongkun Hao$^{2}$ \myand Jing Pan$^{2}$~ Xunying Liu$^{1}$~ Jinyu Li$^{2}$~ Sunit Sivasankaran$^{2}$~ Linquan Liu$^{2}$~ Furu Wei$^{2}$ \\
  $^{1}$The Chinese University of Hong Kong \\
  $^{2}$Microsoft Corporation~~~ \\
  \texttt{sjhu@se.cuhk.edu.hk}~
}

\begin{document}
\maketitle
\begin{abstract}
Recent advancements in large language models (LLMs) have expanded their scope in natural language processing (NLP) to encompass multimodal functions. However, integrating listening capabilities effectively remains a significant challenge for generalization and complex auditory task execution.
In this work, we introduce WavLLM, a robust and adaptive speech large language model featuring dual encoders—a Whisper encoder for semantics and a WavLM encoder for speaker characteristics. 
Within the two-stage curriculum learning framework, WavLLM first builds its foundational capabilities by optimizing on mixed elementary single tasks, followed by advanced multi-task training on more complex tasks combining elementary ones. 
To enhance the flexibility and adherence to different tasks and instructions, a prompt-aware LoRA weight adapter is introduced in the second advanced multi-task training stage.
We validate the proposed model on universal speech benchmarks and also apply it to specialized speech-question-answer (SQA) dataset, and speech Chain-of-Thought (CoT) evaluation set. 
Experiments demonstrate that the proposed model achieves state-of-the-art performance across a range of speech tasks in the setting of the same model size, exhibiting robust generalization capabilities in executing complex tasks using CoT approach. The codes, models, audio samples, and SQA evaluation set can be accessed at \url{https://github.com/microsoft/SpeechT5/tree/main/WavLLM}.
\end{abstract}

\section{Introduction}
Large language models (LLMs) have witnessed a meteoric rise in advancement within the last couple of years, reaching or even exceeding the proficiency of humans in a myriad of natural language processing (NLP) tasks \citep{OpenAI2023GPT4TR, touvron2023llama, anil2023palm}.
With large language models attaining substantial breakthroughs, the focus is increasingly shifting towards the capabilities and advancements of multi-modal large language models (MLLMs), which possess the ability to listen \citep{tang2023salmonn, deshmukh2023pengi}, speak \citep{rubenstein2023audiopalm,hao2023boosting}, see \citep{huang2023language,OpenAI2023GPT4TR}, and create content \citep{pan2023generating,videoworldsimulators2024}.

Amidst the broadening scope of abilities, speech stands out as a crucial form of human communication, prompting extensive research to equip large language models (LLMs) with speech perception capabilities \citep{shu2023llasm,wu2023decoder,wang2023blsp,tang2023salmonn,chu2023qwen, ma2024embarrassingly}.
Typically, LLMs are augmented with an auxiliary audio encoder designed to preprocess audio signals, transforming them into the same input space as that of the LLMs, enabling them to achieve various speech tasks, such as automatic speech recognition (ASR), speech question answering (SQA), and so on. 
However, previous research has yet to overcome significant challenges in achieving effective generalization due to two main issues: (1) specialized tasks are highly sensitive to prompt design, resulting in performance degradation when confronted with unseen or complex instructions; (2) there is an absence of speech Chain-of-Thought (CoT) \cite{wei2022chain} capability, which is essential for addressing complex tasks.


In this work, we propose a robust and adaptive speech large language model, WavLLM, aiming at enhancing the generalization capabilies, following speech instruction effectively, and processing the given speech in accordance with provided textual prompts, as well as supporting multi-round dialog.
Specifically, to distinguish various types of speech information, we utilize a Whisper \citep{radford2023robust} encoder to encode the semantic content of the speech, and a WavLM \citep{chen2022wavlm} encoder to capture the acoustic information, like unique characteristics of the speaker's identity.
\par
During the model training phase, we develop a curriculum learning method that progressively fine-tune LLMs to follow instructions for understanding and processing speech, starting from simple tasks and advancing towards more complex ones.
In the initial mixed single-task training stage, we leverage a substantial dataset of synthesized spoken question-answering content generated by GPT-4 and tailored to various speech-centric tasks such as automatic speech recognition (ASR), speech-to-text translation (ST), emotion recognition (ER), speaker verification (SV), and so on, to fine-tune the WavLLM with Low Rank Adaptation (LoRA) techniques \citep{hu2022lora}.
To enhance the generalization on the unseen or complex instructions\footnote{Please find detailed motivations in Section 2.}, we introduce an advanced multi-task training stage, incorporating a specially constructed prompt-aware multi-task speech processing dataset combining the elementary tasks. Furthermore, we design a novel prompt-aware LoRA weight adapter for this stage, capable of adaptively tuning the LoRA weights according to varying prompts, thereby improving the model's generalization capacity and robustness.

We evaluate the proposed model on \textbf{1) single tasks}, including a) universal speech benchmark, including ASR, SV, ER and ST; b) spoken-query-based question answering and English Listening Comprehension test in Chinese National College Entrance Examination, which presents a spoken dialogue, and requires to answer text-based choice questions related to the conversation, and \textbf{2) multiple tasks}, consisting of c) instruction-independent multi-tasks dataset that combines multiple independent prompts in a single instruction; d) speech CoT evaluation set that decomposes a complex task into multiple sub-tasks.
Extensive evaluations demonstrate that our proposed model exhibits robust generalization and CoT  capabilities, consistently surpassing strong baselines across a broad spectrum of speech-related tasks.

In summary, the contributions of this paper can be categorized as follows:

\noindent 1) Equipped with the proposed prompt-aware LoRA weight adapter,  a curriculum learning method is leveraged for model training, by incrementally fine-tuning large language models with robustness and generalization capabilities, beginning with simple tasks and progressing to complex ones.
\par
\noindent 2) Our proposed model employs a decoupling strategy for speech information, utilizing the Whisper encoder to capture semantic content and the WavLM encoder for acoustic features, thereby enriching speech representation and improving performance on downstream tasks.
\par
\noindent 3) WavLLM demonstrates exceptional generalization capabilities when responding to diverse prompts and completing complex tasks. It exhibits impressive capabilities in zero-shot SQA such as English listening comprehension, and shows strong proficiency in CoT-based tasks, delivering performance gains over non-CoT tasks.

\section{Related Work}
\vspace{-0.2cm}
The exploration of multi-modal large language models involves the integration of diverse data types including text, images, video, speech, audio, and more. This represents a natural progression from text-based large language models, designed to enable the perception of the world and the creation of content \citep{OpenAI2023GPT4TR,huang2023language,hao2023boosting}.
For instance, Kosmos-1 \citep{huang2023language} and GPT-4V \citep{OpenAI2023GPT4TR} are able to perceive general modalities beyond text, and follow instruction provided by users to process and analyze image inputs.
Another research direction focuses on improving the multi-modal generative abilities of language models, enabling them to produce visual content like images or videos, as exemplified by MiniGPT-5 \citep{zheng2023minigpt} and Sora \citep{videoworldsimulators2024}.
Related research to this work focuses on speech-enhanced large language models that aim to endow LLMs with the capability to perceive and process speech signal \citep{zhang2023speechgpt,shu2023llasm,wu2023decoder,tang2023salmonn,chu2023qwen,wang2023blsp}.

Among these studies, SpeechGPT \citep{zhang2023speechgpt} empowers large language models with cross-modal conversational abilities by three-stage training stages, using hidden units as the discrete representation of speech.
LLaSM \citep{shu2023llasm} builds a large Chinese/English speech language model that can understand and follow instructions, through pre-training and cross-modal instruction fine-tuning stages.
BLSP \cite{wang2023blsp} bootstraps Language-Speech Pre-training via behavior alignment of continuation writing.
SALMONN \citep{tang2023salmonn}, named from a speech audio language music open neural network, boosts large language models with generic hearing abilities with a activation tuning stage by playing with the LoRA scaling factor.
Qwen-audio \citep{chu2023qwen} scales up audio-language pre-training to cover over 30 tasks and various audio types, including human speech, natural sounds, music, and songs.

\paragraph{Motivation}
Previous research on Speech Large Language Models (Speech LLMs) has primarily concentrated on executing a single speech task in response to a given instruction, while the feasibility of using a single instruction to simultaneously complete multiple and complex speech tasks has remained unexplored.
The employment of multi-task instructions allows for the efficient completion of several tasks at once and improves performance by dividing complex tasks into logical, related sub-tasks, such as CoT tasks. Such capabilities also suggest the robustness and generalizability of the Speech LLM.
\par
Our initial experiments indicate that (1) prior open-source speech LLMs underperformed in multi-task scenarios, demonstrating these models' limited ability to generalize to complex instructions; (2) reducing the LoRA scaling factor can be beneficial for multi-task instructions, but leads to a substantial degradation of the results of training tasks \citep{tang2023salmonn}, which suggests that single and multiple tasks might benefit from distinct LoRA scaling factors; (3) there is a notable decline in performance when the model encounters unseen or diverse prompts as opposed to seen prompts (3.5\% vs. 2.1\%, see Section \ref{robustness_analysis}), when employing various prompts to evaluate the ASR performance of the open-source model.
Consequently, we introduce a curriculum learning approach that progresses from simple to complex instruction tasks, propose a prompt-aware LoRA weight adapter which dynamically adjusts the amplitude of the LoRA output according to the instruction, and further enhance the generalization by utilizing a diverse array of prompts generated by GPT-4 across all training tasks.

\section{Method}
The WavLLM is optimized by maximizing the following probability:
\begin{equation}
    \small
    p(\bm Y|[\bm X, \bm T]; \bm \Theta) = \prod_{t=0}^{T_Y}p(\bm y_t|[\bm X, \bm T, \bm Y_{<t}]; \bm \Theta)
\end{equation}
where $\bm X$ and $\bm T$ are the speech input and text prompt respectively. $\bm Y = [\bm y_1, \bm y_2, ..., \bm y_{T_Y}]$ is the target text output. $\bm \Theta$ denotes the parameters of WavLLM. The detailed template of WavLLM's training data can be found in Appendix D. 

\begin{figure*}[t]
	\centering
        \includegraphics[width=13cm]{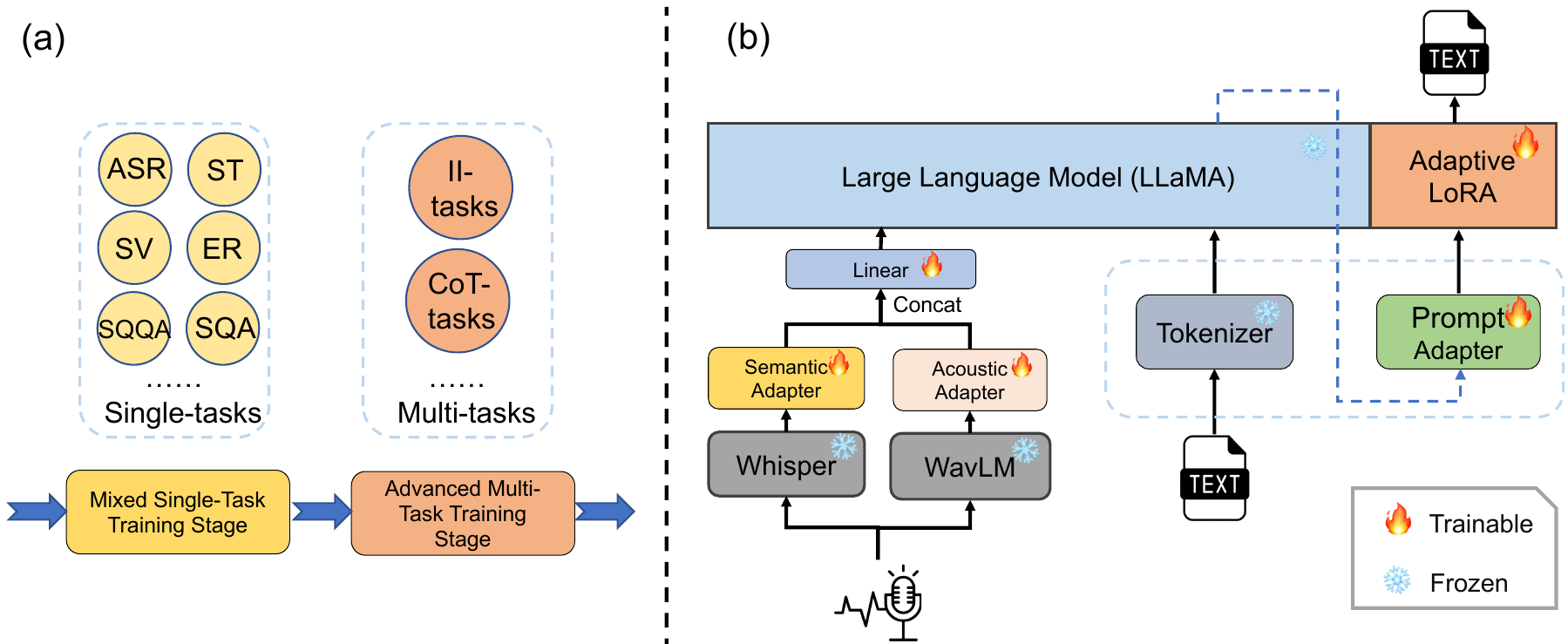}
		\label{fig:stage_2}
    \caption{Overview of the proposed WavLLM. The left part (a) is the two-stage curriculum learning. The right part (b) is the model architecture. Two speech encoders and adapters with different focuses are utilized, where Whisper is used for extracting semantic information, and WavLM for extracting acoustic information. Before being fed to the LLM, these two representations are concatenated together and linearly transformed. Adaptive LoRA approach is used for cross-modal efficient fine-tuning with online adaptation, where the prompt adapter is able to generate prompt-dependent parameters to adjust the amplitude of LoRA in the second advanced multi-task training stage.
    }
	\label{fig:model}
 \vspace{-0.4cm}
\end{figure*}

\subsection{Model Architecture}\label{subsec_3.1}
The model architecture of our framework is shown in Figure \ref{fig:model}, which consists of speech encoders (i.e., Whisper \citep{radford2023robust} and WavLM \citep{chen2022wavlm}) as well as modality adapters, a large language model (i.e., LLaMA \citep{touvron2023llama}) and a proposed prompt adapter.

\paragraph{Speech Encoders and Modality Adapters}
In order to extract both the semantic and acoustic information in the speech, we utilize two state-of-the-art speech encoders, namely Whisper and WavLM. Whisper is trained for ASR and ST tasks in a weakly supervised fashion on a massive 680k-hour speech corpus recorded in diverse conditions, making it well suited for encoding semantic information in speech. WavLM is a predictive based self-supervised learning (SSL) pre-trained model. During its pre-training stage, WavLM mixes each utterance with signals from multiple speakers in the same batch, yet selectively predicts only the targets associated with the utterance's original speaker. Such training method allows WavLM to better extract speaker-related acoustic information. In our work, the 32-layer transformer-based encoder of Whisper-large-v2 and WavLM-base are utilized. Both modality adapters have three components, including two 1-D convolution layers to down-sample and align the output of both encoders within the temporal domain, a down-up bottleneck adapter \citep{houlsby2019parameter}, and a final linear projector. 
The semantic adapter receives its input from the Whisper encoder's output, while the acoustic adapter takes a weighted sum of the hidden states from all layers of WavLM, where the weights are learnable. The outputs of both adapters are concatenated together at the dimension level before feedforwarding into the linear projector.

\paragraph{LLM, LoRA and Prompt Adapter}
Our framework utilizes the LLaMA-2-7B-chat as the LLM backbone, featuring a 32-layer Transformer decoder with an attention dimension of 4096, specifically optimized for dialogue-related use cases.
To integrate the speech modality within the LLM, we employ the parameter-efficient fine-tuning method known as LoRA, which is specifically applied to the key, query, value, and output weight matrices within the attention module of the LLaMA. 

To enable adaptive LoRA scaling factors for different single-task and multiple-task instructions, inspired by adapter layer in \cite{houlsby2019parameter}, we propose an online adaptation strategy by introducing a down-up prompt-aware LoRA weight adapter (aka. prompt adapter) with attention weights, designed to modulate the effect of LoRA on LLaMA, as shown in Figure \ref{fig:model}. 
Given the text-based prompts $\bm T$ with length $M$, we can get the representation $\bm t = [\bm t_1, ..., \bm t_i, ..., \bm t_M] \in \mathbb{R}^{D\times M}$ with LLaMA, where $D$ is the dimension of LLaMA hidden states and $\bm{t} = f(\bm{T}; \bm{\Theta_{\text{LLaMA}}})$. This representation is fed into the prompt adapter to get the LoRA scaling factors, $\bm{r} = [r_1, ..., r_i, ..., r_D] \in  \mathbb{R}^{D \times 1}$:
\begin{align}
    \bm o_i &= \bm P^u \text{GeLU}(\bm P^d \bm t_i) \\
    \bm W &= [\bm{W_A}\bm{o_1}, ..., \bm{W_A}\bm{o_i}, ..., \bm{W_A}\bm{o_M}] \\
    \hat{\bm W} &= \text{Softmax}(\bm W) = [\hat{w_1}, ..., \hat{w_i},..., \hat{w_M}] \\
    r_j &= g(\bm{t_{\cdot j}}; \bm{\Theta_{\text{prompt\_adapter}}}) = \sum_{i=1}^{M}\hat{w_i}\bm{o_{ij}}
\label{equation1}
\end{align}
where $\bm P^u \in \mathbb{R}^{D\times K}$ and $\bm P^d \in \mathbb{R}^{K\times D}$ are up-linear projection and down-linear projection layers respectively, and \text{GeLU} is the GeLU activation function \cite{hendrycks2016gaussian}.  $\bm o_i \in \mathbb{R}^{D\times 1}, \bm W_A\in \mathbb{R}^{1\times D}$ and $\bm{W_A}\bm{o_i} \in \mathbb{R}$ is a scalar. Softmax is applied to $\bm W$ along the sequence length dimension $M$ to obtain the normalized weights. 
The hidden states of an attention layer equipped with adaptive LoRA are expressed by:
\begin{equation}
    \bm{h_i} = \bm{W_0}\bm{x_i} + (\bm{BAx_i}) \odot \bm{r}
\label{equation2}
\end{equation}
where $\bm x = [\bm{x_1}, ... \bm{x_i}, ..., \bm{x_{N+M}}]\in \mathbb{R}^{D\times (N+M)}$ is the input of the attention layer from the speech input $\bm X$ with the length $N$ and text prompt $\bm T$. $\bm{B}\in \mathbb{R}^{D\times R}$ and $\bm A \in \mathbb{R}^{R\times D}$ are the LoRA parameters, $\bm{W_0}\in \mathbb{R}^{D\times D}$ is a weight matrix in the attention layer.

\subsection{Curriculum Learning}\label{subsec_3.2}
In this section, we present the two-stage curriculum-learning (CL) based training method, which facilitates a progression from learning simple data to understanding complex data, thereby enhancing the model's capacity for generalization.

\subsubsection{Mixed Single-Task Training Stage}

In the first stage, various single-task, cross-modal, speech-text pairing datasets or text-only datasets are utilized to endow the LLM with robust capabilities in speech processing and comprehension.
We freeze the parameters of LLM, WavLM, and Whisper encoder, and optimize the modality adapters, a linear layer and LoRA components.

\paragraph{Data Construction}
The first mixed single-task training stage involves various speech tasks, including automatic speech recognition (ASR), speech-to-text translation (ST), speaker verification (SV), emotion recognition (ER), spoken-based instruction tuning and text-based instruction tuning (IT) tasks, as well as a large mount of GPT-generated speech question answering (SQA). There are various questions within the SQA tasks, including those related to the speaker and gender, as well as continuation and summary tasks. Concurrently, these tasks draw upon multiple datasets, including LibriSpeech \citep{Panayotov2015Librispeech} with English reading speech, AMI \citep{carletta2005ami} with multi-talker meeting recordings, as well as Fisher \citep{cieri2004fisher} and Switchboard \citep{godfrey1992switchboard} corpora with 2-channel English telephony conversations. Examples of the training data and prompts used to generate data with GPT-4 can be found in the Appendix A.1 and A.3 respectively. The speech audio clips of spoken-based instruction tuning task are generated by using Microsoft Azure text-to-speech API\footnote{https://azure.microsoft.com/en-us/products/ai-services/text-to-speech}. 
The detailed task information about description, data source, and data hours can be found in Appendix F.

\subsubsection{Advanced Multi-Task Training Stage}

Owing to the incongruity between textual and spoken modalities, extensively fine-tuning the model using the LoRA method on a large amount of prompt-repetitive speech-text data, such as ASR and ST tasks, may cause the model to overfit on specific speech tasks, thereby compromising the LLM's powerful instruction-following capabilities. For instance, the model exhibits subpar performance when handling multi-task instructions, often only managing to accomplish a fraction of the tasks assigned.
Specifically, if ASR is included in the tasks, the model might complete only the ASR portion while failing to address the remaining instructions.

To this end, we construct a more complex prompt-aware multi-task dataset in the second stage, by integrating various single-task instructions. Multi-task and single-task datasets are utilized together in this training stage.
Besides, we noted that simply incorporating more challenging training data may slightly diminish the performance of single-task instructions, such as ASR, when compared to results of the first training phase. Hence we introduce a prompt adapter, as illustrated in Section \ref{subsec_3.1}, to produce adaptive LoRA scaling factors for different instructions and tasks, and serve as an effective approach to concurrently enhance the model's generalization capabilities.

\paragraph{Data Construction} Given a speech audio clip, we combine different task prompts for this audio segment as well as text-based instruction tuning tasks together as instructions. The training target is designed to complete the tasks sequentially and to repeat key parts of each prompt prior to delivering a response.  For example, for an utterance in LibriSpeech, ASR, SQA and text-based IT (t-IT) tasks can be combined into multi-task dataset. Please refer to Appendix A.2 for specific examples. In our work, a total of 2.9K hours of various multitask data are used, including ER+t-IT, ASR+t-IT, ST+t-IT, SV+t-IT, SQA+t-IT, ASR+ST, ASR+SQA, ASR+ST+t-IT and ASR+SQA+t-IT combined tasks, which are summarized in Appendix F.

\section{Experiments}
Please find implementation details in Appendix G.

\begin{table}
\centering
\caption{Single-task and multi-task evaluation benchmarks, including tasks, datasets, and metrics. ``Acc.'' stands for accuracy.}
\setlength\tabcolsep{1.5pt}
\scalebox{0.65}{
\begin{tabular}{c|c|c|c|c} 
\hline\hline
\multicolumn{2}{c|}{Task}                            & Dataset                      & Split                                                                                       & Metric                     \\ 
\hline\hline
\multirow{10}{*}{\begin{tabular}[c]{@{}c@{}}Single\\-task\end{tabular}} & \multirow{2}{*}{ASR} & \multirow{2}{*}{LibriSpeech} & test-clean                                                                                  & \multirow{2}{*}{WER (\%)}  \\
                              &                      &                              & test-others                                                                                 &                            \\ 
\cline{2-5}
                              & \multirow{2}{*}{ST}  & \multirow{1}{*}{CoVoST2 \cite{wang2020covost}}     & \multirow{2}{*}{En2De}                                                                                       & \multirow{2}{*}{BLEU}      \\

                              &                      & \multirow{1}{*}{MUSTC \cite{di2019must}}       &                                                                                        &                            \\
\cline{2-5}
                              & SV                   & VoxCeleb1 \cite{nagrani2017voxceleb}                    & test set                                                                                    & Acc. (\%)              \\ 
\cline{2-5}
                              & ER                   & IEMOCAP \cite{busso2008iemocap}                      & Session 5                                                                                   & Acc. (\%)              \\ 
\cline{2-5}
                              & SQQA                 & WikiQA \citep{yang-etal-2015-wikiqa}                       & test set                                                                                    & Acc. (\%)              \\ 
\cline{2-5}
                              & SQA                  & MuTual \citep{cui2020mutual}                       & test set & Acc. (\%)              \\ 
\hline\hline
\multirow{3}{*}{\begin{tabular}[c]{@{}c@{}}Multi\\-task\end{tabular}}   & II-Task           & In-house, based on MuTual                     & -                                                                                           & Acc., IFR (\%)                    \\ 
\cline{2-5}
                              & \multirow{2}{*}{CoT}                   &  \begin{tabular}[c]{@{}c@{}}In-house, based on \\Gigaword \citep{graff2003english}\end{tabular}                    & -                                                                                           &  \multirow{2}{*}{\begin{tabular}[c]{@{}c@{}}R-1, R-2, \\ R-L,\\ BERTScore \end{tabular}}                         \\
\cline{3-4}
                              &                  &  \begin{tabular}[c]{@{}c@{}}In-house, based on \\story generated by GPT-4 \end{tabular}                    & -                                                                                           &                    \\
\hline\hline
\end{tabular}
}
\label{table_2}
\vspace{-0.4cm}
\end{table} 

\begin{table*}[htbp]
\centering
\caption{Single-task instruction performance of our WavLLM model compared to other open-source speech large language models and cascaded Whisper+LLM baseline model. ``*'' stands for the results from public paper.}
\vspace{-0.2cm}
\scalebox{0.75}{
\begin{tabular}{c|cc|cc|c|c|c|c} 
\hline\hline
\multirow{3}{*}{Models} & \multicolumn{2}{c|}{ASR}              & \multicolumn{2}{c|}{ST (En2De)}       & \multirow{2}{*}{SV} & \multirow{2}{*}{ER} & \multirow{2}{*}{SQQA} & \multirow{2}{*}{SQA}  \\ 
\cline{2-5}
                        & test-clean & test-other               & CoVoST2  & MUSTC                      &                     &                     &                       &                       \\ 
\cline{2-9}
                        & \multicolumn{2}{c|}{WER$^\downarrow$} & \multicolumn{2}{c|}{BLEU$^\uparrow$~} & Acc.$^\uparrow$     & Acc.$^\uparrow$     & Acc.$^\uparrow$       & Acc.$^\uparrow$       \\ 
\hline\hline
Whisper + LLM           & 2.7$^*$    & 5.2$^*$                  & 18.2     & 11.5                       & -                   & -                   & 0.78                  & 59.30\% (63.50\%)     \\ 
\hline
SALMONN-7B              & 2.4        & 5.4                      & 17.1     & 12.5                       & 0.86                & -                   & -                     & 39.95\% (40.00\%)     \\
SALMONN-13B             & 2.1$^*$    & 4.9$^*$                  & 18.6$^*$ & 19.5                       & 0.94$^*$            & 0.69$^*$            & 0.41$^*$              & 43.35\% (43.35\%)     \\ 
\hline
Qwen-Audio-Chat 7B      & 2.2        & 5.1                      & 23.2     & 18.4                       & 0.50                & -                   & 0.38                  & 25.50\% (54.25\%)     \\ 
\hline
Our WavLLM 7B           & \textbf{2.0}        & \textbf{4.8}                      & \textbf{23.6}     & \textbf{21.7}                       & \textbf{0.91}                & \textbf{0.72}                & 0.57                  & \textbf{67.55\% (67.55\%)}     \\
\hline\hline
\end{tabular}
}
\label{main_table_single}
\vspace{-0.4cm}
\end{table*}

\subsection{Evaluation Setup}
\label{evaluation_setup}
Corresponding to the training methods, two primary levels of testing tasks were evaluated, namely, single-task and multi-task evaluations. 
The detailed information of the two types of task evaluations are provided in the Table \ref{table_2}. 
Single-task evaluation consists of ASR, ST, SV, ER, SQA, and spoken-query-based question answering (SQQA). 
The main difference between SQQA and SQA is that in SQQA the questions are directly in the audio, whereas in SQA the questions are given by text-based instructions.
In our work, the single-answer multiple-choice questions of English Listening Comprehension examination (Gaokao) in China are used as the zero-shot SQA task, which gives a short dialogue, a question, and three options. The model is required to choose the correct one from three options. The performance of SQA is not only a measure of the comprehension of the cross-modal speech and textual content, but also serves as an indicator of the model's generalization capabilities with respect to a diverse array of instructions.

In the multi-task evaluation, two distinct types of tasks are tested, both of which are given a speech audio clip: the tasks that consist of independent instructions (II-Task) and the tasks that feature sequentially progressive instructions, which are also known as CoT tasks. Examples of these two tasks can be found in the Appendix B. For II-Task, our focus lies on not only the ability to follow instructions, i.e. instruction following rate (IFR)\footnote{The IFR is scored manually on 10\% of the test utterances.}, but also the correct completion of each instruction. Whereas for CoT tasks, our primary concern is the performance of the final instruction, which will be compared to the performance of one-shot non-CoT based instructions.
The multitasking instruction of zero-shot II-tasks includes ASR, SQA, ST and the general knowledge question task. 
The audio for zero-shot CoT task is generated from the Gigaword \citep{graff2003english} dataset using Microsoft Azure text-to-speech API, and the target German texts are translated from English summaries of Gigaword dataset\footnote{Translation directions of ASR+SQA+ST tasks in second advanced training stage are all English to Chinese.} by utilizing GPT-4.
The CoT task requires the Speech LLM to complete ASR, summary and translation tasks in turn.
In contract, the one-shot non-CoT based instructions require the cross-lingual summarization directly.
For open-ended or target-lack test sets, GPT-4 is utilized to score the outputs, including the accuracy of SQQA and II-task, which is conducted three times and then take the average to minimize the randomness from GPT-4.

\begin{table*}[htbp]
\centering
\caption{Multi-task instruction performance of our WavLLM model compared to other open-source speech LLMs.}
\vspace{-0.2cm}
\scalebox{0.75}{
\begin{tabular}{c|cc|cccc|cccc} 
\hline\hline
\multirow{2}{*}{Models} & \multicolumn{2}{c|}{II-tasks} & \multicolumn{4}{c|}{CoT (ASR+SUMM+En2De, gigaword)} & \multicolumn{4}{c}{w/o CoT (De\_SUMM, gigaword)}  \\ 
\cline{2-11}
                        & Acc.$^\uparrow$  & IFR$^\uparrow$                  & R-1$^\uparrow$  & R-2$^\uparrow$ & R-L$^\uparrow$  & BERTScore$^\uparrow$             & R-1$^\uparrow$  & R-2$^\uparrow$ & R-L$^\uparrow$  & BERTScore$^\uparrow$          \\ 
\hline\hline
SALMONN-7B              & 22.49 & 34.50                 & 11.9 & 2.4 & 10.7 & 66.46                 & 15.0 & 3.3 & 13.5 & 69.50               \\ 
SALMONN-13B             & 19.58 & 24.25                 & 10.9 & 2.1 & 9.8  & 68.12                 & 14.0 & 2.9 & 12.6 & 69.11               \\
\hline
Qwen-Audio-Chat 7B         & 37.99 & 57.75                 & 5.9  & 0.9 & 5.7  & 67.62                 & 5.8  & 0.9 & 5.3  & 65.84               \\ 
\hline
Our WavLLM 7B        & \textbf{62.44} & \textbf{92.50}                 & \textbf{16.5} & \textbf{4.1} & \textbf{14.7} & \textbf{70.60}                 & \textbf{15.4} & \textbf{3.8} & \textbf{13.9} & \textbf{70.37}               \\
\hline\hline
\end{tabular}
}
\label{main_table_multi}
\vspace{-0.4cm}
\end{table*}

\subsection{Main Results}
We compare the performance of WavLLM with other open source text-instruction (chat) based speech LLMs, including SALMONN \citep{tang2023salmonn} and Qwen-Audio-Chat \citep{chu2023qwen}, as well as the baseline system that cascades Whisper large-v2 with LLaMA-2-7b-chat, across various single-task and multi-task instructions.
\paragraph{Single-task Evaluation}
As shown in Table \ref{main_table_single}, for the ASR task, our chat model achieves state-of-the-art WERs of 2.0\% and 4.8\% on test-clean and test-other sets of LibriSpeech corpus, surpassing other open-source chat models on the same size (7B). Similar superior performance are observed in ST, SV, ER and SQQA tasks.
\par
The SQA task in our paper is the zero-shot English listening comprehension tests. As shown in column ``SQA'' of Table \ref{main_table_single}, two types of accuracy are evaluated: a) the correct option must be explicitly given (the first number); b) answers that are semantically equivalent to the correct option is considered correct (the second number), which are scored by GPT-4 (The scoring instruction can be found in Appendix C.1). 
The larger the both accuracy, the better the model's comprehension and generalization capacity, while the smaller the difference between the both accuracy, the better the model's ability to follow instructions. From the results, we can observe that our WavLLM model consistently surpasses the cascaded Whisper + LLaMA baseline, and other open source speech LLMs (67.55\% vs. 25.50-59.30\%).
Additionally, our WavLLM model supports multiple dialogue scenario, with a representative instance detailed in Appendix E.

\paragraph{Multi-task Evaluation} As shown in Table \ref{main_table_multi}, 
despite the optimization of SALMONN through activation tuning, and the fact that Qwen-Audio-Chat conducts fine-tuning only on audio encoder without impacting LLM by LoRA weights, their performance in following multitasking instructions remains significantly suboptimal.
Our final chat model produces a markedly higher instruction-following rate for the II-Task compared to SALMONN and Qwen-Audio-Chat (92.50\% vs. 24.25\%-57.75\%), which suggests the necessity and effectiveness of our advanced multi-task training stage with prompt adapter. 
From the accuracy based on GPT-4, which further focuses on whether they are completed correctly, similar trend can be observed (62.44\% vs. 19.58\%-37.99\%). The scoring instruction can be found the Appendix C.2. 
\par
When the model is able to handle multi-task instructions, we aspire to enhance its performance by Chain of Thought (CoT) methodology. Specifically, the CoT based prompt is excepted to give a better performance than one-shot non-CoT based instructions. We list the examples of these two types of prompts in the Appendix B.2. From the results in Table \ref{main_table_multi}, we can draw two conclusions: 1) Our WavLLM model produces the best performance on the CoT-task instructions; 2) Compared with the performance given one-shot non-CoT instructions, our model produces consistent performance improvements on all metrics.


\begin{table}[htbp]
\vspace{-0.1cm}
\centering
\caption{Model performance with/without advanced training on multi-task instructions. \textit{mixed training} and \textit{advanced training} stand for the first and training stage. ``BS.'' refers to BERTScore \cite{zhang2019bertscore}. $\dagger$ denotes statistically significant improvement obtained over the model of mixed training.}
\setlength\tabcolsep{1.5pt}
\scalebox{0.6}{
\begin{tabular}{l|cc|cccc|cccc} 
\hline\hline
\multirow{3}{*}{\hspace{8mm}Models}      & \multicolumn{2}{c|}{\multirow{2}{*}{II-tasks}} & \multicolumn{8}{c}{CoT (ASR+SUMM+En2De)}                                                                                                           \\ 
\cline{4-11}
                             & \multicolumn{2}{c|}{}                          & \multicolumn{4}{c|}{gigaword}                                           & \multicolumn{4}{c}{story}                                                \\ 
\cline{2-11}
                             & Acc.$^\uparrow$ & IFR$^\uparrow$               & R-1$^\uparrow$ & R-2$^\uparrow$ & R-L$^\uparrow$ & BS.$^\uparrow$ & R-1$^\uparrow$ & R-2$^\uparrow$ & R-L$^\uparrow$ & BS.$^\uparrow$  \\ 
\hline\hline
\textit{mixed training}      & 22.92           & 26.25                        & 14.7           & 3.3            & 13.2           & 69.71                & 18.0           & 2.9            & 13.7           & 68.61                 \\
+ \textit{advanced training} & 62.44           & 92.50                        & 16.5$^\dagger$           & 4.1$^\dagger$            & 14.7$^\dagger$           & 70.60                & 24.5$^\dagger$           & 4.8$^\dagger$            & 19.0$^\dagger$           & 72.52                 \\
\hline\hline
\end{tabular}
}
\label{advanced_training_multi}
\end{table}

\vspace{-0.5cm}
\subsection{Analysis}
\paragraph{The Effect of Advanced Training}
Table \ref{advanced_training_multi}\footnote{Significant tests for ASR, ST, SV, ER, SQA and CoT tasks (other tasks are scored by GPT-4) are performed. For ASR task, a matched pairs sentence-segment word error (MAPSSWE \cite{gillick1989some}) based statistical significance test at a significance level of 0.05 is performed. And refer to \cite{dror2018hitchhiker}, for ST and CoT tasks (R-1, R-2, R-L), paired bootstrap resampling \cite{koehn2004statistical} at a significance level of 0.05 is performed. For SV, ER and SQA, t-test at a significance level of 0.05 is performed.} shows the results of our models after first mixed single-task training stage and second advanced multi-task training stage\footnote{The results of single-task instructions can be found in Appendix H. After advanced training, our model produces comparable or even better performance on single-task prompts compared to the first-stage model.}. 
For zero-shot II-tasks, significant enhancement of generalization ability is obtained after advanced training, as evidenced not only by the increased adherence to instructions (92.50\% vs. 26.25\%) but also by the higher accuracy of each executed instruction (62.44\% vs. 22.92\%). For cross-lingual summary tasks using CoT based instructions, our advanced multi-task trained model consistently outperforms the first stage model. 
\par
In addition, we found that the first stage model mainly accomplished the ST task and did not perform the summarization task. To better demonstrate the effectiveness of the second stage, we crafted a long story-based CoT task by GPT-4 where the audio contains a 100-word story, and the target is a 20-word summary in German.
In this task, if the model solely focuses on completing the translation, there will be a noticeable discrepancy in length between its output and the target. From the results of this task in Table \ref{advanced_training_multi}, the second advanced multi-task training stage model significantly outperforms the first stage model, up to 65.52\% relative improvement on R-2. When compared to SALMONN-7B on story-based CoT task instructions, a similar greater enhancements can be obtained (24.5/4.8/19.0/72.52 vs. 10.6/1.3/7.8/63.90 on R-1, R-2, R-L and BERTScore respectively.).

\begin{figure*}[htbp]
	\centering
	\begin{subfigure}{0.24\linewidth}
		\centering
		\includegraphics[width=\linewidth]{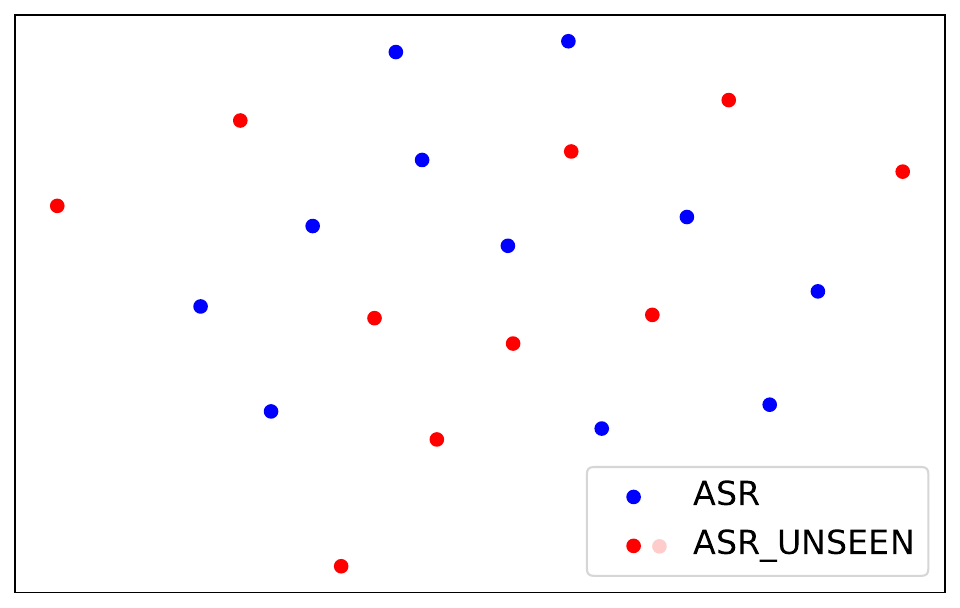}
		\label{fig:asr_sv_emo_st_sum_asr+st}
        
	\end{subfigure}
	\centering
	\begin{subfigure}{0.24\linewidth}
		\centering
		\includegraphics[width=\linewidth]{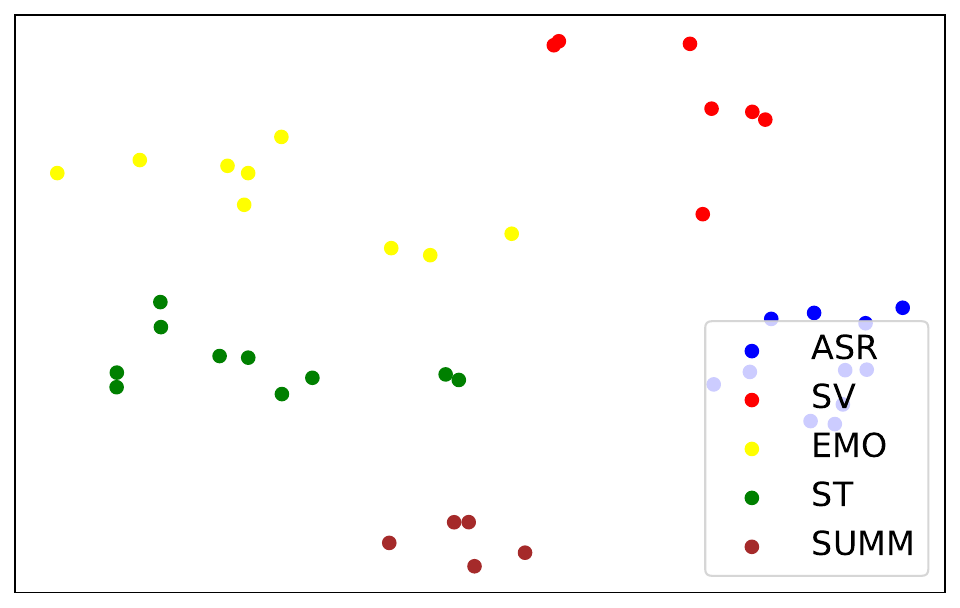}
		\label{fig:asr_asr_unseen}

	\end{subfigure}
	\label{fig:methods}
    \centering
    \begin{subfigure}{0.24\linewidth}
		\centering
		\includegraphics[width=\linewidth]{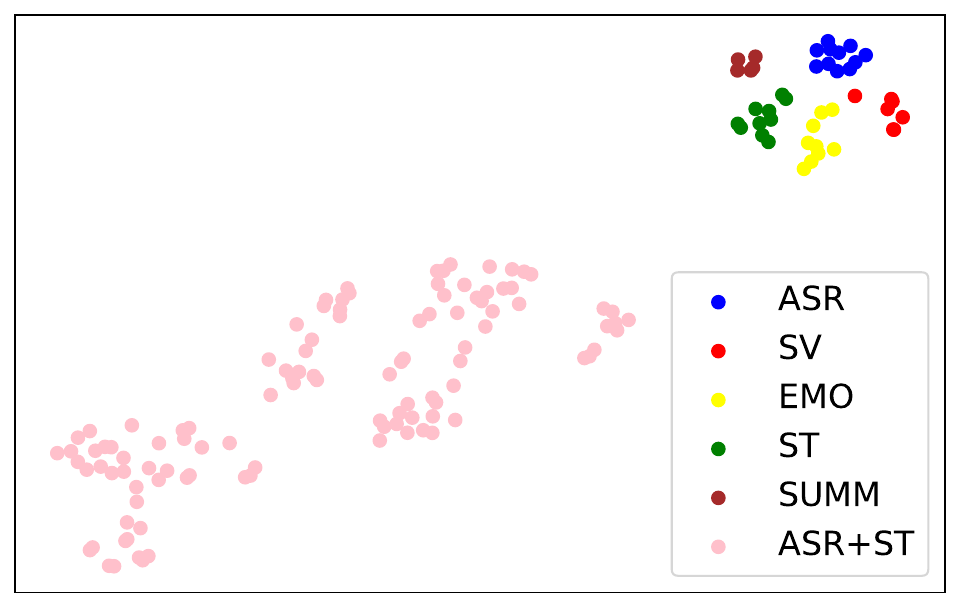}
	\label{fig:asr_sv_emo_st_sum_asr+st}
        
	\end{subfigure}
	\centering
	\begin{subfigure}{0.24\linewidth}
		\centering
		\includegraphics[width=\linewidth]{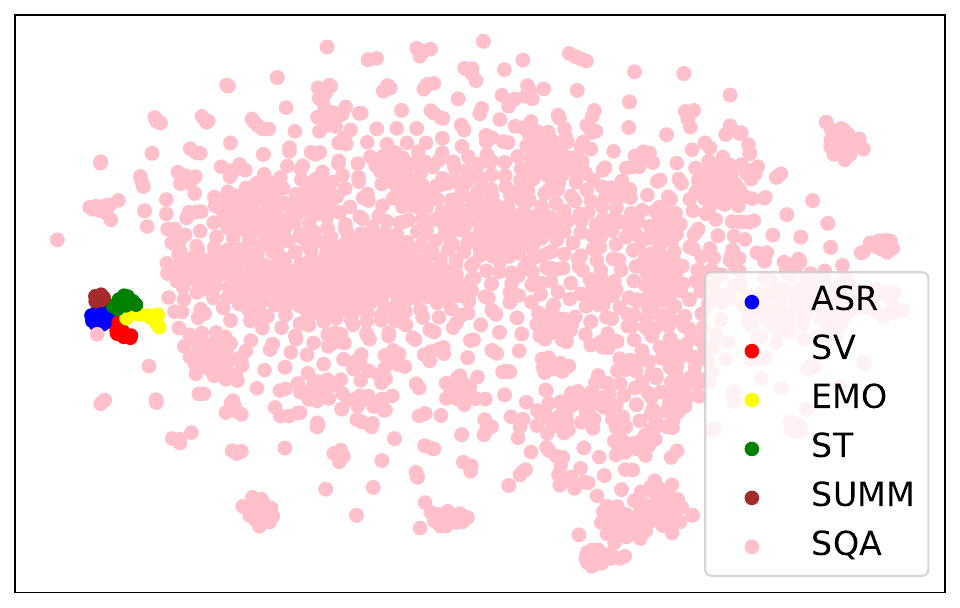}
		\label{fig:asr_asr_unseen}

	\end{subfigure}
    \vspace{-0.6cm}
    \caption{TSNE visualization of the proposed prompt adapter's outputs. Each point corresponds to a prompt.
    }
	\label{fig:tsne}
 \vspace{-0.6cm}
\end{figure*}

\paragraph{The Effect of Prompt Adapter}

\begin{table}[htbp]
\vspace{-0.3cm}
\centering
\caption{Model performance across training stages with/without a prompt adapter on single-task instructions. \textit{one-stage} denotes the model is trained by utilizing all single-task and multi-task data simultaneously. \textit{two-stage (LoRA)} stands for two-stage training method with only LoRA technique. ``t-c'', ``t-o'', ``CoV.'' and ``MU.'' stand for test-clean, test-other, CoVoST2 and MUSTC. $\dagger$ denotes statistically significant improvement of the model with the prompt adapter over the model of two-stage (LoRA); $\ast$ denotes statistically significant improvement of the model of two-stage (LoRA) over the model of one-stage.}
\vspace{-0.1cm}
\setlength\tabcolsep{2pt}
\scalebox{0.7}{
\begin{tabular}{l|cc|cc|c|c|c|c} 
\hline\hline
\multirow{3}{*}{\hspace{8mm}Models}   & \multicolumn{2}{c|}{ASR}              & \multicolumn{2}{c|}{ST (En2De)}       & \multirow{2}{*}{SV} & \multirow{2}{*}{ER} & \multirow{2}{*}{SQQA} & \multirow{2}{*}{SQA}  \\ 
\cline{2-5}
                          & t-c & t-o             & CoV. & MU.                      &                     &                     &                       &                       \\ 
\cline{2-9}
                          & \multicolumn{2}{c|}{WER$^\downarrow$} & \multicolumn{2}{c|}{BLEU$^\uparrow$~} & Acc.$^\uparrow$     & Acc.$^\uparrow$     & Acc.$^\uparrow$       & Acc.$^\uparrow$       \\ 
\hline\hline
\textit{one-stage}        & 2.1         & 5.0                     & 22.7     & 21.0                       & 0.88                & 0.71                & 0.51                  & 65.35\%               \\
\textit{two-stage (LoRA)} & 2.1         & 5.1                     & 23.3$^\ast$     & 21.2                       & 0.89$^\ast$                & 0.71                & 0.54                  & 63.70\%               \\
+ Prompt Adapter          & 2.0         & 4.9$^\dagger$                     & 23.6$^\dagger$     & 21.6                       & 0.90$^\dagger$                & 0.72                & 0.57                  & 65.00\%               \\
\hline\hline
\end{tabular}
}
\label{stage_prompt_adapter_single}
\vspace{-0.3cm}
\end{table}

Despite the fact that data-level curricular learning benefits the performance on complex cross-modal tasks, using the same LoRA parameters between single-task and multi-task instructions may diminish the performance on both instructions.
A prompt-aware LoRA weight adapter (prompt adapter) is proposed to address this issue. Comparative experiments are conducted to analyze the effect of prompt adapter during the second advanced multi-task training stage.
Additionally, we build a one-stage model trained by combining all data, including both single-task and multi-task data\footnote{Due to the computing resource constraints, only a portion of the single-task dataset are utilized during the second advanced multi-task training stage in this section.}. 

From the results of Table \ref{stage_prompt_adapter_single} and \ref{stage_prompt_adapter_multi}, the following conclusions can be drawn.
Firstly, the results of two-stage model without a prompt adapter against one-stage model further demonstrate that the two-stage curriculum learning based training is effective as evidenced by 1) the comparable performance of single-task instructions; 2) consistent performance improvements on zero-shot II-task and CoT-task prompts.
Secondly, incorporating the proposed prompt adapter consistently outperforms the baseline two-stage model without such module on all single-task and multi-task instructions.

\begin{table}[htbp]
\centering
\caption{Model performance across training stages with/without a prompt adapter on multi-task prompts.}
\scalebox{0.7}{
\begin{tabular}{l|cc|cccc}
\hline\hline
\multirow{2}{*}{\hspace{8mm}Models} & \multicolumn{2}{c|}{II-tasks}  & \multicolumn{4}{c}{CoT (gigaword)}  \\ 
\cline{2-7}
                        & Acc.$^\uparrow$                   & IFR$^\uparrow$   & R-1$^\uparrow$  & R-2$^\uparrow$ & R-L$^\uparrow$  & BS.$^\uparrow$             \\ 
\hline\hline
\textit{one-stage}               & 59.34 & 85.50 & 14.8 & 3.4 & 13.2 & 69.64                 \\
\textit{two-stage (LoRA)}        & 61.15                  & 90.25 & 15.8$^\ast$ & 3.8$^\ast$ & 14.5$^\ast$ & 70.42                 \\ + Prompt Adapter        & 63.05                  & 92.75 & 16.5$^\dagger$ & 4.0 & 14.8 & 70.75                 \\
\hline\hline
\end{tabular}
}
\label{stage_prompt_adapter_multi}
\vspace{-0.3cm}
\end{table}

\vspace{-0.2cm}
\paragraph{The Effect of WavLM}
WavLM model has been widely used for speech processing as a foundation model, especially for speaker information extraction. Table \ref{w_wavlm} shows the single-task instruction performance on models with or without WavLM encoder after the first mixed single-task training stage. Incorporating the weighted sum of all layers in WavLM-base encoder not only brings performance improvements to speaker verification task but also enhances other tasks such as ASR (relative WER reductions of 13.04\% and 11.11\% on test-clean and test-other) and ST tasks.

\begin{table}[htbp]
\vspace{-0.1cm}
\centering
\caption{Single-task instruction performance of models w or w/o WavLM encoder after the mixed training. $\dagger$ denotes statistically significant improvement obtained over WavLLM w/o WavLM.}
\setlength\tabcolsep{2pt}
\vspace{-0.1cm}
\scalebox{0.65}{

\begin{tabular}{c|cc|cc|c|c|c|c} 
\hline\hline
\multirow{3}{*}{Models} & \multicolumn{2}{c|}{ASR}              & \multicolumn{2}{c|}{ST (En2De)}      & \multirow{2}{*}{SV} & \multirow{2}{*}{ER} & \multirow{2}{*}{SQQA} & \multirow{2}{*}{SQA}  \\ 
\cline{2-5}
                        & t-c & t-o             & CoV. & MU.                      &                     &                     &                       &                       \\ 
\cline{2-9}
                        & \multicolumn{2}{c|}{WER$^\downarrow$} & \multicolumn{2}{c|}{BLEU$^\uparrow$} & Acc.$^\uparrow$     & Acc.$^\uparrow$     & Acc.$^\uparrow$       & Acc.$^\uparrow$       \\ 
\hline\hline
WavLLM                  & 2.0$^\dagger$         & 4.8$^\dagger$                     & 23.9$^\dagger$    & 21.9$^\dagger$                       & 0.91$^\dagger$                & 0.72                & 0.55                  & 67.30\%               \\
WavLLM w/o WavLM        & 2.3         & 5.4                     & 23.4    & 21.0                       & 0.89                & 0.73                & 0.55                  & 68.55\%               \\
\hline\hline
\end{tabular}
}
\label{w_wavlm}
\vspace{-0.3cm}
\end{table}

\vspace{-0.2cm}
\paragraph{Robustness Analysis} \label{robustness_analysis}
In this subsection, the robustness\footnote{In this paper, ``robustness'' refers to the model's ability to maintain stable performance on in-domain data across different conditions, such as varying prompts, including seen and unseen prompts; On the other hand, the ``generalizability'' measured by the model's performance on out-of-domain data, e.g., the performance of zero-shot SQA task.} of the speech LLMs is evaluated by comparing the performance between the seen and the unseen prompts on our WavLLM model and SALMONN model\footnote{Various prompts generated by GPT-4 are used as unseen prompts.}. From the results in Table \ref{seen_unseen}, compared to the SALMONN model, which experienced a decline in performance with unseen or diverse prompts, our WavLLM model does not exhibit any performance degradation with unseen prompts on ASR tasks and even produces performance improvement on the ST task, demonstrating our model's powerful robustness.

\begin{table}[htbp]
\vspace{0.1cm}
\centering
\setlength\tabcolsep{2.5pt}
\caption{
Model performance using seen(se.) or unseen (unse.) prompts on WavLLM and SALMONN.
}
\label{seen_unseen}
\vspace{-0.2cm}
\scalebox{0.75}{
\begin{tabular}{c|cc|cc|cc} 
\hline\hline
\multirow{3}{*}{Models} & \multicolumn{4}{c|}{ASR (WER$^\downarrow$)}                                            & \multicolumn{2}{c}{ST-CoVoST2 (BLEU$^\uparrow$)}       \\ 
\cline{2-7}
                        & \multicolumn{2}{c|}{test-clean} & \multicolumn{2}{c|}{test-other} & \multicolumn{2}{c}{En2De}            \\ 
\cline{2-7}
                        & se. & unse.              & se. & unse.              & se. & unse.                  \\ 
\hline\hline
SALMONN-7B              & 2.4     & 81.8                   & 5.4     &  85.5                      & 17.1    & 15.9                       \\
SALMONN-13B             & 2.1     & 3.5                    & 4.9     &  8.8                      & 18.6    & 18.2                       \\ 
\hline
Our WavLLM 7B           & 2.0     & 2.0                    & 4.8     & 4.8                    & 23.4    & 23.6                       \\
\hline\hline
\end{tabular}
}
\vspace{-0.3cm}
\end{table}

\paragraph{Visualization of LoRA Weights}
In this subsection, TSNE \citep{van2008visualizing} based visualization of the proposed prompt adapter's output is given in Figure \ref{fig:tsne}. Several trends can be observed: 1) The overlap between two clusters of the seen and unseen ASR prompts implies the generalization of the proposed prompt adapter; 2) the clear discrimination among single-task prompts suggests that the proposed prompt adapter is capable of discerning various single-task instructions and assigning disparate weights to each; 3) Similar strong discrimination between single-task and multi-task instructions is obtained which validates our motivation; 4) The wide distribution of the SQA task with various prompts, illustrates that the prompt adapter can accommodate diverse prompts.

\section{Conclusion}
\label{conclusion}
In this paper, we propose WavLLM, a robust and adaptive speech large language model, which uses LLaMA-2-chat as foundational LLM backbone, and extracts semantic and acoustic information from speech audio utilizing Whisper and WavLM encoders. 
Utilizing a curriculum learning approach, the proposed WavLLM commences with single-task instruction training in the initial mixed training phase and subsequently expanding our training to encompass additional multi-task prompts in the second advanced phase with the integration of the proposed prompt adapter.
Massive experiments demonstrate that our WavLLM model delivers state-of-the-art performance on various speech-related tasks and robust generalization abilities on single-task and multi-task instructions, enhanced by a CoT processing ability that greatly improves its effectiveness in tackling intricate tasks.


\section*{Limitations}
Although WavLLM shows a remarkable proficiency in handling speech-related tasks and impressive cross-modal instruction following and generalization capacity, it also exhibits some specific constraints.
\paragraph{Adaptive Use of CoT} Our WavLLM model produces performance improvements using CoT based instructions compared to one-shot non-CoT based instructions. However, it lacks the capability to autonomously decompose complex one-shot non-CoT based tasks into CoT based ones. For future work, we are interested in advancing the capability of adaptive use of CoT. This requires WavLLM to determine whether a task can be decomposed into multiple sub-tasks, and then applying the CoT approach accordingly.
\paragraph{Broader Applicability} Although our WavLLM model focuses primarily on speech in English, it can be readily extended to accommodate multiple languages.
Additionally, the WavLLM model excels at processing and comprehending spoken language, yet it lacks the capability to generate speech. We defer the task of expanding WavLLM's capabilities to synthesize speech to future research.
\paragraph{Safety and Ethics} Employing continuous speech representations within our WavLLM model may render it more vulnerable to adversarial attacks, potentially undermining the model's compliance with the HHN criteria (Harmless, Helpful, Honest). This vulnerability merits further investigation for solutions.

\bibliography{my_bib}
\clearpage
\input{appendix}

\end{document}

%% file: appendix.tex
\appendix

\section{Some Examples of Training Data}
\subsection{SQA Task}\label{App_A_1}
\begin{tabular}
{p{6cm}|p{3cm}|p{2cm}|p{1.5cm}}
\hline\hline
speech transcript                                                & instructions                                                                                               & targets                                                                              & source       \\ 
\hline
the nautilus would still have nothing to fear from their attacks & What animal would still have nothing to fear from attacks? & The nautilus. & LibriSpeech \\
\hline
as gelatinous or glutinous matters impress the sense of touch i have spoken both of sound and of voice i mean to say that the sound was one of distinct of even wonderfully thrillingly distinct & What is the gender of the speaker? And what is the speaker talking about? & The speaker is female. And she is talking about sound and voice being distinct. & LibriSpeech \\
\hline
A: so how do you feel about it; B: uh i don't see any problem with with random- testing for random drugs; A: yeah i myself almost um am in favor of it uh; A: i work at Honeywell and i went through a a pretty i don't know i went through a standard drug testing thing before i was brought on; B: um-hum & What is the second speaker's opinion on random drug testing? & The second speaker doesn't see any problem with random drug testing. & Switchboard \\
\hline\hline
\end{tabular}
\clearpage

\subsection{Multi-task Instruction Datasets}\label{App_A_2}
\par
\begin{tabular}{p{2cm}|p{4cm}|p{5cm}|p{1.5cm}}
\hline\hline
speech transcript                                                & instructions                                                                                               & targets                                                                              & source       \\ 
\hline
the nautilus would still have nothing to fear from their attacks & Initially, Analyze the speech in the given audio clip and provide a complete textual transcription; Then, answer the question: What animal would still have nothing to fear from attacks?; Thirdly, ignore the audio clip, Come up with 5 activities for a birthday party for a 5-year-old. & The audio transcription is: the nautilus would still have nothing to fear from their attacks \textbackslash{}n For the question ``{}What animal would still have nothing to fear from attacks?'', the response is: The nautilus. \textbackslash{}n The response to ``{}Come up with 5 activities for a birthday party for a 5-year-old. '' is: 1. Play a game of freeze dance \textbackslash{}n 2. Make balloon animals \textbackslash{}n 3. Have a color scavenger hunt \textbackslash{}n 4. Bake cupcakes  \textbackslash{}n 5. Make a craft project together & LibriSpeech, Alpaca \\
\hline\hline
\end{tabular}

\clearpage

\subsection{The Prompts for Generating SQA Data by GPT-4}\label{App_A_3}

\begin{tabular}{p{13cm}}
\hline\hline
You are asked to generate **only one** questions, and their corresponding answers, according to some conversational sentences given below. These sentences have been transcribed from conversational speech data with one or multiple speakers who are taking to each other. ``Speaker A'' and ``Speaker B'' in the senteces are labeled by human and your response must not contain human-marked information, namely ``Speaker A'' and ``Speaker B''. Here are the requirements: 1. Your response should strictly follow the format below: {``Question'': ``xxx'', ``Answer'': ``xxx''}; 2. Please ignore ``Speaker A'' and ``Speaker B'' in the given sentences. Your response should strictly not include the phrase ``Speaker A'' and ``Speaker B''; 3. Your question should be highly related to the conversation, and your answer must be **correct**, and should be simple and clear. Besides, you question should be designed as your answer has to be reasoned from the conversation; 4. For example, a sentence ``Speaker A: It is a good day; Speaker B: Yes, but I have to go to hospital'' means that speaker A first say it is a good day and speaker B then say that Yes, but I have to go to hospital. 5. **Very Importance**: Your questions and answers **can not** contain the word ``Speaker A'' and ``Speaker B'', because ``Speaker A'' and ``Speaker B'' in the sentences are additional labels for transcripts, and they are different people. For example, the question ``What is Speaker B's opinion?'' **does not** meet the requirements because it contains word ``Speaker B''. Namely, you can not use ``Speaker A'' and ``Speaker B'' to represent they in questions and answers, maybe you can use the first or second speaker to denote ``Speaker A'' or ``Speaker B''; 6. The type of response should be diverse. The respone **must contain** double quotation marks for each part. Here are the sentences: {transcript}
\\
\hline\hline
\end{tabular}
\clearpage

\section{Some Examples of Evaluation Data}\label{App_B}

\subsection{Examples of II-task Instruction}\label{App_B_1}
\par
\begin{tabular}{p{4cm}|p{6cm}|p{1cm}|p{1cm}}
\hline\hline
speech transcript                                                & instructions                                                                                               & targets                                                                              & source       \\ 
\hline
Women: ``How much time do you usually spend exercising daily?'' 
Man: ``Frankly speaking, I'm an awfully lazy man. I know it's time to change.'' & To begin, What will the man do next? A. Start to take exercise; B. Do as he always does; C. Change his working time.; Next, Create a French transcript for this English audio clip; Furthermore, Recognize the speech and give me the transcription; Last step, setting aside the audio, Who wrote ``The Adventures of Sherlock Holmes''? & - & MuTual  \\
\hline\hline
\end{tabular}

\subsection{Examples of CoT-task and Non-CoT-task Instruction}\label{App_B_2}
\par
\begin{tabular}{p{5cm}|p{4cm}|p{2cm}|p{1cm}}
\hline\hline
speech transcript                                                & instructions                                                                                               & targets                                                                              & source       \\ 
\hline
three films from Asia-Pacific are in the running for the coveted golden palms at this year\'s Cannes film festival, competing in a field dominated by European productions, organizers announced Monday.   &  First of all, transcribe the audio recording into text, capturing every spoken word; Additionally given this audio clip and text, can you condense it into a clear, concise summary, no more than 20 words?; Lastly disregarding the sound, translate this English summary into German. & Drei Filme aus dem asiatisch-pazifischen Raum im Rennen in Cannes & gigaword \\
\hline
three films from Asia-Pacific are in the running for the coveted golden palms at this year\'s Cannes film festival, competing in a field dominated by European productions, organizers announced Monday.   &  Please summarize the content of the audio clip in German, no more than 20 words. & Drei Filme aus dem asiatisch-pazifischen Raum im Rennen in Cannes & gigaword \\
\hline\hline
\\
\end{tabular}

\clearpage

\section{The Prompt for Scoring using GPT-4}
\subsection{SQA Scoring}\label{App_C_1}
\begin{tabular}{p{13cm}}
\hline\hline
Next, I will give you a multiple-choice question along with its correct answer, as well as a generated answer that needs to be evaluated for correctness. You will need to determine whether the given answer is correct based on the question and the correct answer, and give a simple reason. The answer must explicitly give the correct option to be considered correct and not by implication or indirect response. Your response should strictly follow the format:\{"result": "xx", "reason": "xx"\}, if the given answer is correct, then your response should be \{"result": "True", "reason": "xx"\}, otherwise your response should be \{"result": "False", "reason": "xx"\}.Here is the question: \{"What will the man do next? A. Start to take exercise; B. Do as he always does; C. Change his working time."\},and the correct answer is \{"A"\},the answer that needs to be judged is \{"B. Do as he always does"\}.
\\
\hline\hline
\end{tabular}

\clearpage

\subsection{II-task Scoring}\label{App_C_2}
\begin{tabular}{p{15cm}}
\hline\hline
Next i will give you an audio transcription, instructions related or unrelated to the audio, and the corresponding responses. You need to use the given information to figure out how many instructions were completed correctly in the given responses and how many were left unanswered or answered incorrectly, then give the simple and clear reason why each question was completed or not, Finally, you need to do the self-examination to ensure responses you give must be correct and without inconsistencies.\\You must adhere to the following \textbf{rules}: 1. instructions are only answered sequentially, and one answer must only correspond to one instruction; 2. For choice questions, the answer must be in **English** and only **one** of the three options, multiple options or none of the options are considered incomplete, e.g. the answer is ``A. xx; B. xx; C. xx.'' (note that options may appear across sentences or lines); 3. All instructions must be completed in **English**, except translation tasks; 4. The translation task is not a translation of instructions, but a translation of the content of the given audio speech 5. If the response is a repetition of the instruction, then it is considered not completed.\\Your response should strictly follow the format: \{``instructions completed'': ``xx'', ``instructions incompleted'': ``xx'', ``Reasons for each instructions'': [``instruction\_1: xx'', ``instruction\_2: xx'', ``instruction\_3: xx'', ``instruction\_4: xx'']\}.
\\
\hline
You need to refer to \textbf{this example}: the transcription of the audio: ``So James What are you going to do for your birthday? Well, I was hoping to have a party but most of my friends are busy so I have accepted my sisters offer to take me to a film. At least it is better than going out for a meal with my parents.''; the instructions are ``To begin, ignore the audio clip, Who is known for developing the theory of relativity?; Then, What will the man do on his birthday? A. Have a party; B. See a movie; C. Go out for a meal.; Moreover, Convert the spoken words in this audio file into a textual format; Last step, Translate this English speech into Japanese";\\and the respones are ``1. \begin{CJK*}{UTF8}{min}
エイトリングの理論を発展した人物は誰ですか？
\end{CJK*}; 2. What will the man do on his birthday? A. Have a party; B. See a movie; C. Go out for a meal.; 3. \begin{CJK*}{UTF8}{min}この音声>ファイルをテキスト形式に変換してください。\end{CJK*}; 4.  \begin{CJK*}{UTF8}{min}この英語のスピーチを日本語に翻訳してください。\end{CJK*}''.\\And your output should be \{``instructions completed'': ``0'', ``instructions incompleted'': ``4'', ``Reasons for each instructions'': [``instruction\_1: incompleted, the answer is the Japanese translation of the instruction. Q\&A task should be completed in English.'', ``instruction\_2: incompleted, the answer is the a repetition of the instruction. More than one options were given'', ``instruction\_3: incompleted, the answer is the Japanese translation of the instruction'', ``instruction\_4: incompleted, the answer is the Japanese translation of the instruction'']\}.\\Be sure to carefully follow the rules and refer to the examples for responses.\\
\hline
\textbf{Here is the transcription of the audio}: \{``I really want to go to the concert, but I haven't got a ticket yet leave it to me, I'll organize that we can go together really that's wonderful.''\},\\ \textbf{the instructions} are \{``To begin, disregarding the sound, How many elements are on the periodic table?; Then, Translate this English speech into Japanese; Additionally, How will the speaker get a ticket to the concert? A. The man will go to buy the ticket; B. The woman will get the ticket; C. The man will have someone buy the ticket.; Lastly, Transcribe the given audio clip into text''\},\\\textbf{the responses} is \{``Addressing the prompt 'How many elements are on the periodic table?', the answer is: There are currently 118 elements on the periodic table. From the audio, we have the following translations: \begin{CJK*}{UTF8}{min}コンサートに行きたいけど まだチケットがないの どうするんだ？\end{CJK*} For the question ``How will the speaker get a ticket to the concert? A. The man will go to buy the ticket; B. The woman will get the ticket; C. The man will have someone buy the ticket.'', the response is: B; The audio transcription is: i really want to go to the concert but i haven't got a ticket yet b  leave it to me i'll organize that we can go together r  really that's wonderful \}
\\
\hline\hline
\end{tabular}

\clearpage
\section{Prompt Template for WavLLM}\label{App_D}
The template of our WavLLM containing a speech audio clip, text instructions and targets are constructed as follows: \\
\\
\begin{tabular}{p{8cm}}
\hline
\textless bos\textgreater\lbrack INST\rbrack\textless\textless SYS\textgreater\textgreater\textbackslash n As a helpful language and speech assistant, you are able to understand the speech content provided by the user, and assist the user with a variety of tasks using natural language. \textbackslash n\textless\textless/SYS\textgreater\textgreater\textbackslash n\textbackslash n \textless SPEECH\textgreater `` speech '' \textless/SPEECH\textgreater``instruction'' \lbrack/INST\rbrack`` target ''\textless eos\textgreater \\
\hline
\end{tabular}
\\
\\
where ``speech'' will be replaced by the 4096-dim speech features extracted from speech encoders and modality adapters, while ``instruction'' and ``target'' are the specific task prompts and outputs. The input to the WavLLM is this template with the <eos> removed, while the target is this template without the <bos>. During training, only the ``target'' part is involved in the loss calculation.

\section{Example of Multi-round Dialog}\label{App_E}
\begin{figure}[H]
    \centering
    \includegraphics[width=12cm]{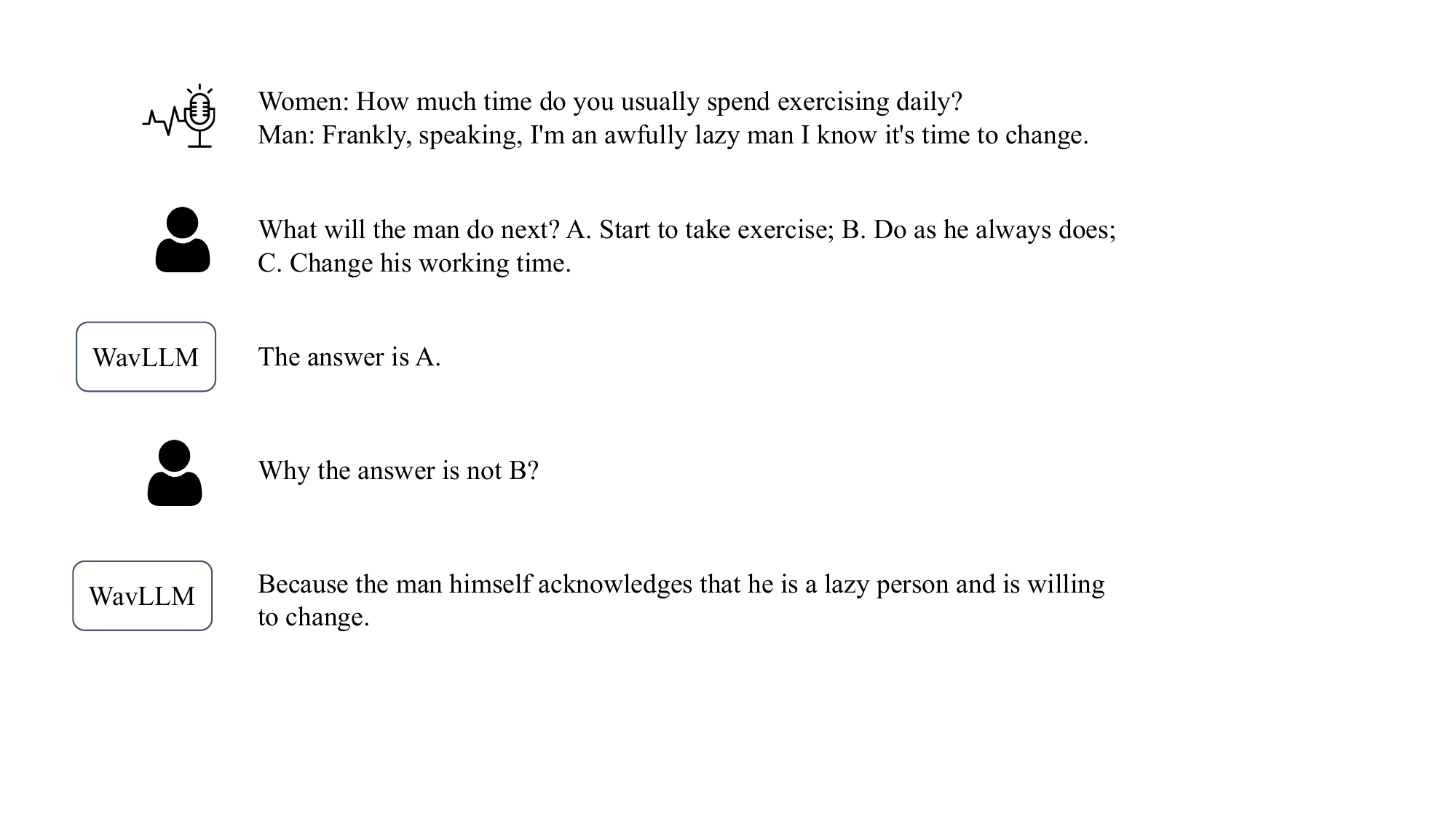}
    \caption{An example of multi-round dialog}
    \label{fig:enter-label}
\end{figure}

\clearpage
\section{Training Data Details}\label{App_F}

Training data used in the first stage and second stage. For all tasks, the instructions are diverse. ``\#Hours'' refers to the duration of speech data for each task, not the total number of hours of the data source. The targets of SQA tasks are generated using GPT3.5, GPT-4 or LLaMA-2-chat.
\par
\vspace{1cm}
\scalebox{0.75}{
\begin{tabular}{c|c|c|c} 
\hline\hline
Task                          & Description                                                                                                                                                                                  & Data Source                                                   & \#Hours  \\ 
\hline\hline
\multirow{13}{*}{\begin{tabular}[c]{@{}c@{}}Single\\-task\end{tabular}} & \multirow{2}{*}{automatic speech recognition (ASR)}                                                                                                                                          & LibriSpeech                                                   & 960      \\
                              &                                                                                                                                                                                              & LibriHeavy medium                                              & 1800     \\ 
\cline{2-4}
                              & \multirow{2}{*}{\begin{tabular}[c]{@{}c@{}}speech-to-text translation (ST), including English to German (En2De), \\English to Japanese (En2Ja), and English to Chinese (En2Zh)\end{tabular}} & CoVoST2                                                        & 440      \\
                              &                                                                                                                                                                                              & MuST-C                                                        & 280      \\ 
\cline{2-4}
                              & speaker verification (SV)                                                                                                                                                                    & VoxCeleb                                                     & 1290     \\ 
\cline{2-4}
                              & emotion recognition (ER)                                                                                                                                                                     & \begin{tabular}[c]{@{}c@{}}IEMOCAP Session 1-4  \end{tabular} & 5        \\ 
\cline{2-4}
                              & \multirow{4}{*}{\begin{tabular}[c]{@{}c@{}}speech question answering (SQA), \\including gender and speaker-related questions, \\ and multi-round QA\end{tabular}}                            & LibriSpeech                                                   & 520      \\
                              &                                                                                                                                                                                              & AMI                                                           & 50       \\
                              &                                                                                                                                                                                              & Fisher                                                         & 710      \\
                              &                                                                                                                                                                                              & Switchboard                                                   & 230      \\ 
\cline{2-4}
                              & speech question answering (SQA), continue writing tasks                                                                                                                                      & LibriSpeech                                                   & 960      \\ 
\cline{2-4}
                              & speech question answering (SQA), summary tasks                                                                                                                                               & LibriSpeech                                                   & 410      \\ 
\cline{2-4}
                              & instruction tuning (IT), including spoken based and text based tasks                                                                                                                         & Alpaca                                                       & 90       \\ 
\hline\hline
\multirow{9}{*}{\begin{tabular}[c]{@{}c@{}}Multi \\-task\end{tabular}}   & ER + text based IT                                                                                                                                                                           & IEMOCAP Session 1-4, Alpaca                                   & 71       \\ 
\cline{2-4}
                              & ASR + text based IT                                                                                                                                                                          & LibriSpeech,~ Alpaca                                          & 274      \\ 
\cline{2-4}
                              & ST + text based IT                                                                                                                                                                           & CoVoST2,~MuST-C, Alpaca                                       & 343      \\ 
\cline{2-4}
                              & SV + text based IT                                                                                                                                                                           & VoxCeleb,~Alpaca                                              & 243      \\ 
\cline{2-4}
                              & SQA + text based IT                                                                                                                                                                          & AMI, Fisher,Switchboard, Alpaca                & 773      \\ 
\cline{2-4}
                              & ASR + ST                                                                                                                                                                                     & LibriSpeech                                                   & 74       \\ 
\cline{2-4}
                              & ASR + SQA                                                                                                                                                                                    & LibriSpeech                                                   & 43       \\ 
\cline{2-4}
                              & ASR + ST + text-based IT                                                                                                                                                                     & CoVoST2, Alpaca                                               & 5        \\ 
\cline{2-4}
                              & ASR + SQA + text-based IT                                                                                                                                                                    & LibriSpeech, Alpaca                      & 1066     \\
\hline\hline
\end{tabular}
}
\clearpage

\section{Implementation Details}\label{App_G}
As mentioned above, the semantic and acoustic speech encoders are the encoder of Whisper-large-v2\footnote{https://huggingface.co/openai/whisper-large-v2} and WavLM-base\footnote{https://huggingface.co/microsoft/wavlm-base}, the backbone LLM is LLaMA-2-chat-7b\footnote{https://huggingface.co/meta-llama/Llama-2-7b-chat-hf}, and all of their parameters are frozen. The outputs of both modality adapters have a time stride of 80 ms and a dimension of 2048, and the rank ($R$) of LoRA is set as 32. In the first mixed single-task training stage, the total number of parameters in our model is 7.55 billion, of which 76.6 million are trainable. In the advanced training phase, the bottleneck dimension ($K$) of the prompt adapter is set as 1024. The 4096-dimensional prompt-dependent parameters produced by prompt adapter are element-wise multiplied with the outputs of the LoRA. Our models are trained with the two-stage curriculum-learning method on 32 V100 GPUs using the Adam optimizer, set with hyperparameters $\beta_1 = 0.9$, $\beta_2 = 0.98$ and batch size equivalent to 30 seconds per GPU, where the first stage consisted of 400K steps and the subsequent stage involved an additional 150K steps. Additionally, we employed a maximum learning rate of $1 \times 10^{-4}$, incorporating a warm-up phase for the first 10\% of steps.
The two-stage training data are presented in data construction part of Section \ref{subsec_3.2}.

\section{The Effect of Advanced Training for Single-tasks}\label{App_H}

Performance of model with or without advanced training on single-task instructions. \textit{mixed training} means the first mixed single-task training stage, and \textit{advanced training} means the second advanced multi-task training stage.

\vspace{1cm}
\scalebox{0.7}{
\begin{tabular}{l|cc|cc|c|c|c|c} 
\hline\hline
\multirow{3}{*}{\hspace{8mm}Models}      & \multicolumn{2}{c|}{ASR}              & \multicolumn{2}{c|}{ST (En2De)}      & \multirow{2}{*}{SV} & \multirow{2}{*}{ER} & \multirow{2}{*}{SQQA} & \multirow{2}{*}{SQA}  \\ 
\cline{2-5}
                             & test-clean & test-other             & CoVoST2 & MUSTC                      &                     &                     &                       &                       \\ 
\cline{2-9}
                             & \multicolumn{2}{c|}{WER$^\downarrow$} & \multicolumn{2}{c|}{BLEU$^\uparrow$} & Acc.$^\uparrow$     & Acc.$^\uparrow$     & Acc.$^\uparrow$       & Acc.$^\uparrow$       \\ 
\hline\hline
\textit{mixed training}      & 2.0         & 4.8                     & 23.9    & 21.9                       & 0.91                & 0.72                & 0.55                  & 67.30\%               \\
+ \textit{advanced training} & 2.0         & 4.8                     & 23.6    & 21.7                       & 0.91                & 0.72                & 0.57                  & 67.55\%               \\
\hline\hline
\end{tabular}
}
\label{advanced_training_single}

\section{The Creation Process of Multi-task Instruction Dataset}\label{App_I}
Take the ``ASR + text based IT'' training task as an example. We use the speech of the ASR task as the input speech, and the instruction is a combination of the respective instructions from the two tasks, linked together using connectives. For example, the instruction of ASR and IT tasks are ``Analyze the speech in the given audio clip and provide a complete textual transcription'' and ``Come up with 5 activities for a birthday party for a 5-year-old.'', then the combined instruction is ``Initially, Analyze the speech in the given audio clip ...; Then, ignore the audio clip, Come up with 5 activities for ...''. The training target is designed to complete the tasks sequentially and to repeat key parts of each prompt prior to delivering a response. In this case, it is ``The audio transcription is: ... \textbackslash n The response to ``Come up with 5 activities for a birthday party for a 5-year-old. '' is: ...''. The creation process of evaluation sets is similar to the training set.
\clearpage
\section{Some Case Study and Error Analysis}\label{App_J}
\par
\begin{tabular}{p{3cm}|p{4cm}|p{7cm}}
\hline\hline
speech transcript                                                & instructions                                                                                               & response                                                 \\ 
\hline
How much time do you usually spend exercising daily? Frankly, speaking, I'm an awfully lazy man I know it's time to change. & To begin, What will the man do next? A. Start to take exercise; B. Do as he always does; C. Change his working time.; \textcolor{red}{Next, Create a French transcript for this English audio clip}; Furthermore, Recognize the speech and give me the transcription; Last step, setting aside the audio, Who wrote 'The Adventures of Sherlock Holmes'? & In reply to 'What will the man do next? A. Start to take exercise; B. Do as he always does; C. Change his working time.', the answer is: A. \textcolor{red}{The audio clip's translation reads: How much time do you usually spend exercising daily? I frankly speaking, I'm an awfully lazy man, I know it's time to change.} The audio clip's transcription reads: how much time do you usually spend exercising daily. As for 'Who wrote 'The Adventures of Sherlock Holmes'?', the answer provided is: Sir Arthur Conan Doyle  \\
\hline
What would you like pizza, sandwich, hotdog, fried rice? we do everything takeaway. fried rice sounds good. OK fried. no on second thought. I'll have a tomato sandwich. & Initially, What will the woman have? A. Fried rice; B. A sandwich; C. Pizza.; Next, Create a French transcript for this English audio clip; Furthermore, setting aside the audio, What is the capital of Germany?; Last step, Please provide a text version of the speech in the provided audio & The answer to 'What will the woman have? A. Fried rice; B. A sandwich; C. Pizza.' is: B. \textcolor{red}{The audio clip's translation reads: QG : Qu'est-ce que tu veux ? A. Pizza. B. Sandwich. C. Hot-dog. D. Frite de riz. A. : Tu veux une pizza ? B. Un sandwich. C. Hot-dog. D. Frite de riz. A. : Tiens, une pizza ? B. Un sandwich. C. Hot-dog. D. Frite de riz. A. Tu veux une pizza ? B. Un sandwich. C. Hot-dog. D. Frite de riz. A. Tu veux une pizza ? B. Un sandwich. C. Hot-dog. D. Frite de riz...}   \\
\hline\hline
\end{tabular}
\vspace{0.3cm}
\par
One of the error cases arises from a failure to distinguish between ``transcription'' and ``translation'', and the other is due to the repetition of the same responses, a problem also observed in other multimodal large language models.